\RequirePackage[2018-12-01]{latexrelease}

\documentclass{tADR2e}

\begin{document}

\jvol{35} \jnum{23} \jyear{2021} \jmonth{Sep}

\title{Estimation of Articulated Angle in Six-Wheeled Dump Trucks using Multiple GNSS Receivers for Autonomous Driving}

\author{Taro Suzuki$^{a}$$^{\ast}$\thanks{$^\ast$Corresponding author. Email: taro@furo.org
\vspace{6pt}},  
Kazunori Ohno$^{b}$, 
Syotaro Kojima$^{b}$, 
Naoto Miyamoto$^{b}$, 
Takahiro Suzuki$^{b}$,
Tomohiro Komatsu$^{c}$, 
Yukinori Shibata$^{d}$, 
Kimitaka Asano$^{e}$, and 
Keiji Nagatani$^{f}$
\\\vspace{6pt} 
$^{a}${\em{Future Robotics Technology Center, Chiba Institute of Technology, Japan}};
$^{b}${\em{New Industry Creation Hatchery Center, Tohoku University, Japan}};
$^{c}${\em{Kowatech Co., Ltd., Japan}};
$^{d}${\em{Sato Koumuten Co., Ltd., Japan}};
$^{e}${\em{Sanyo-Technics Co., Ltd., Japan}};
$^{f}${\em{The University of Tokyo, Japan}};
\\\vspace{6pt}
\received{v1.0 released March 2021} }

\maketitle

\begin{abstract}
  Due to the declining birthrate and aging population, the shortage of labor in the construction industry has become a serious problem, and increasing attention has been paid to automation of construction equipment. We focus on the automatic operation of articulated six-wheel dump trucks at construction sites. For the automatic operation of the dump trucks, it is important to estimate the position and the articulated angle of the dump trucks with high accuracy. In this study, we propose a method for estimating the state of a dump truck by using four global navigation satellite systems (GNSSs) installed on an articulated dump truck and a graph optimization method that utilizes the redundancy of multiple GNSSs. By adding real-time kinematic (RTK)-GNSS constraints and geometric constraints between the four antennas, the proposed method can robustly estimate the position and articulation angle even in environments where GNSS satellites are partially blocked. As a result of evaluating the accuracy of the proposed method through field tests, it was confirmed that the articulated angle could be estimated with an accuracy of 0.1$^\circ$ in an open-sky environment and 0.7$^\circ$ in a mountainous area simulating an elevation angle of 45$^\circ$ where GNSS satellites are blocked.
    
\medskip

\end{abstract}

\section{Introduction}
Currently, the construction industry is facing a serious shortage of labor due to the declining birth rates and aging population, and this is becoming a major problem. In order to solve this labor shortage and realize labor saving in construction work, automation of construction machines is attracting a great deal of attention. Automation of various types of construction machinery is being promoted mainly by construction machinery manufacturers.

\begin{figure}[t]
  \centering
  \includegraphics[width=120mm]{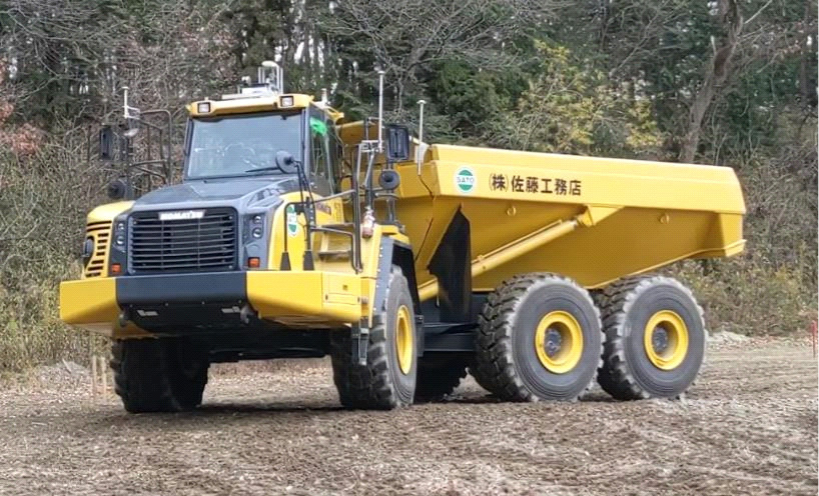} 
  \caption{Six-wheeled articulated dump truck for sediment transportation. The front and rear sections of the dump truck bend to control the direction of travel.}
  \label{fig:1}
\end{figure}

Automatic operation systems for various types of construction machinery have been researched and developed \cite{con1,con2,con3}. Sediment loading and unloading using a dump truck and a backhoe is one of the most basic operations at a construction site \cite{congeneral1,congeneral2}. To automate the sediment loading and unloading, we researched the automatic operation technology of articulated six-wheel dump trucks \cite{our1,our2}. Figure \ref{fig:1} shows an articulated six-wheel dump truck consisting of front and rear sections that contain the dump body. The articulated dump truck differs from a normal dump truck in that it controls the direction of travel by hydraulically turning the front section against the rear section and bending the joint connecting the front and rear. This mechanism is effective for uneven and soft terrains. The angle sustained by this joint is called the "articulated angle.” This articulated angle works like the steering angle of a normal dump truck.

For the automatic operation of dump trucks, it is important to measure the state of the dump truck, such as self-position, orientation, and articulated angle, with high accuracy. In the automatic operation of a dump truck for the sediment loading and unloading, the required positional accuracy is about 0.1 m to stop the dump truck at a pinpoint position. In addition, it is important to estimate the articulated angle to be less than 1 degree for automatic operation. In particular, the articulated angle is very important for automatic steering control of dump trucks. However, in general, it is very difficult to measure the articulated angle because the front and rear sections of the dump truck are physically far apart. The articulate angle changes with steering wheel operation, but the position of the steering wheel and the articulate angle do not coincide. It is assumed that the articulate angle is measured as internal information of the dump truck, but the information is not disclosed. Therefore, it is necessary to install external sensors to measure the articulated angle for automatic operation. In our previous research, we proposed a method to measure the articulated angle using a wire-type linear encoder \cite{our1} and small light detection and ranging (LiDAR) \cite{our2}; however, these methods have disadvantages such as large installation time and less reliability.

In this study, we propose a method to estimate the position and articulated angle of a six-wheeled dump truck with high accuracy by using four global navigation satellite system (GNSS) antennas and receivers installed on a dump truck. We propose a state-estimation method based on graph optimization with various constraints in the GNSS measurements. We use the geometric arrangement of GNSS antennas and GNSS carrier-phase observations to generate various constraints between antennas, which enables the estimation of the articulated angle of a dump truck with a simple installation method.

The contribution of this research is as follows:
\begin{itemize}
  \item This is the first study in which the position and articulated angle of a dump truck are estimated with high accuracy using multiple GNSSs.
  \item By developing an optimization method using the constraints generated by GNSS measurements, highly accurate state estimation is possible even in an environment where the GNSS satellite is blocked, such as in mountainous areas.
\end{itemize}

\section{Related Works}
Various methods for the automatic operation of construction machinery have been developed, mainly by the manufacturers \cite{congeneral1,congeneral2}. To make an existing dump truck operate automatically, it is required to operate the steering wheel, accelerator pedal, and brake pedal of the dump truck in some way. In some studies, dump trucks are directly operated by controlling their hydraulic system \cite{art1,art2}. However, this mechanism of operation is difficult for users because construction machinery manufacturers do not disclose vehicle information. Therefore, in this study, a robot was retrofitted to physically control the dump truck. We developed a robot named SAM, which enables the remote control of an existing hydraulic dump truck \cite{our1}. Using the SAM, we can control the target articulated dump truck.

For the automatic operation, the state estimation of the dump truck is essential. A method of retrofitting GNSS has been proposed to estimate the status of construction equipment, such as normal dump trucks and backhoe \cite{dumpgnss1}. This method uses real-time kinematic (RTK)-GNSS to estimate the position of the construction machinery with centimeter accuracy, and it is possible to estimate the orientation using two GNSS antennas \cite{twognss1}. However, actual construction sites are often located in mountainous areas, and in such environments where GNSS satellites are blocked, the high-precision fix solution of RTK-GNSS cannot be obtained, and the performance of a position and orientation estimation is degraded. There are some examples of automatic operation of normal dump trucks equipped with GNSS and inertial navigation system (INS) \cite{twognss2}, However, normal GNSS/INS cannot measure the articulated angle of an articulated dump truck, which is physically divided into front and rear sections, and cannot be directly applied to the automatic operation of a six-wheeled dump truck.

Nonetheless, there are few examples of studies about state estimation for articulated dump trucks. For articulated dump trucks, it is necessary to measure the articulated angle, in addition to the position and orientation. As mentioned above, methods to measure the articulated angle using a wire-type encoder or LiDAR have been proposed. Two wire-type encoders were used to measure the length between the front and rear sections of the articulated dump truck \cite{our1}. The articulated angle can be estimated from the change in the distance measured by the two wire-type encoders. However, wire-type encoders are difficult to maintain and have reliability problems for use in harsh construction sites. LiDAR was used to measure the distance instead of the wire-type encoder \cite{our2}. However, LiDAR requires a complex calibration process for mounting the position and orientation. These methods also have many problems, such as high maintenance costs.

In this study, we propose retrofitting four GNSS receivers/antennas to estimate the articulated angle of a six-wheeled dump truck. The proposed method is easy to install and greatly improves the performance and convenience as compared to the previous methods. We propose a method for estimating the state of a dump truck using a graph-based optimization technique. Graph-based optimization has been actively studied in the field of robotics and is widely used for optimization-based state estimation problems, such as simultaneous localization and mapping (SLAM) \cite{slam1,slam2,slam3}. Graph-based optimization consists of nodes corresponding to various states and edges representing constraints between nodes. Although there are several studies on using GNSS for graph-based optimization \cite{gognss0,gognss1,gognss2}, the method of using multiple GNSSs has not been studied so far.

\section{Proposed Method}
\subsection{Configuration of GNSS antennas on dump truck}
\begin{figure}[t]
  \centering
  \includegraphics[width=150mm]{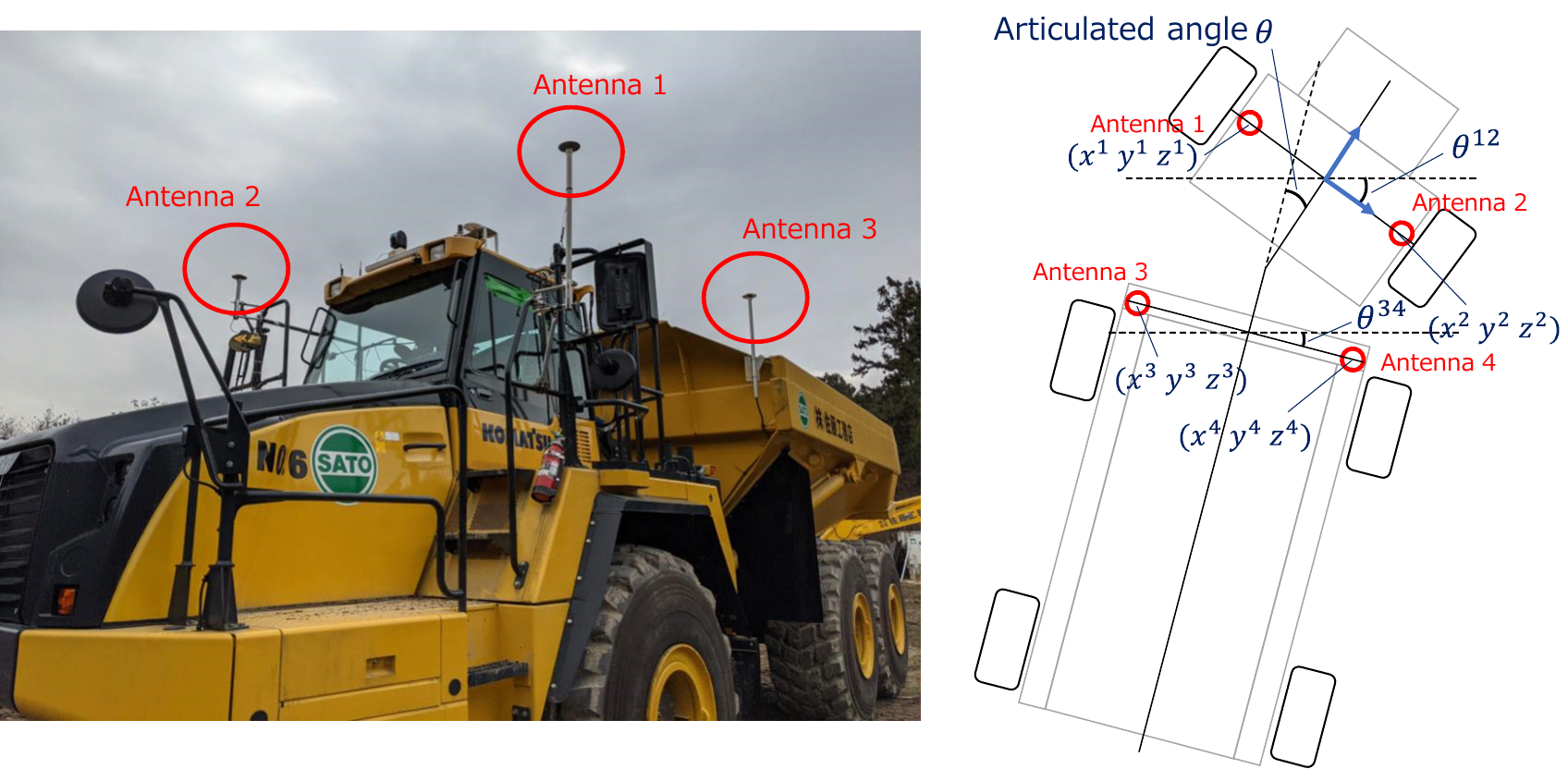} 
  \caption{Four GNSS antennas were installed on the front and rear sections of a dump truck. The articulated angle can be calculated from the positions of the four GNSS antennas. }
  \label{fig:2}
\end{figure}

Figure \ref{fig:2} shows the appearance of the dump truck equipped with four GNSS antennas, the configuration of the GNSS antennas, and the coordinate system of the dump truck. As shown in Figure \ref{fig:2}, two GNSS antennas are installed in the front section and in the rear section of the articulated dump truck. The GNSS antennas are attached to the pole with magnets and can be easily removed. Based on the geometric arrangement of the four GNSSs, the position and articulated angle of the dump truck shown in Figure \ref{fig:2} were estimated. Here, the articulated angle is defined as the angle between the front and rear of the dump truck projected onto the horizontal plane.

\subsection{Flow of Prosed Method}
\begin{figure}[t]
  \centering
  \includegraphics[width=120mm]{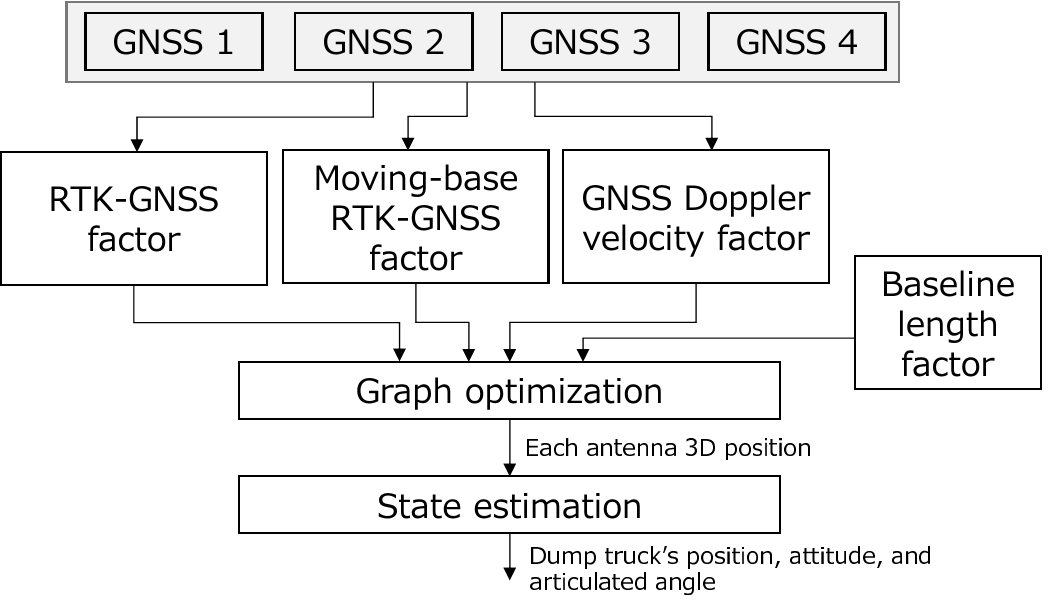} 
  \caption{Flow of proposed method. We use graph-based optimization technique with four factors.}
  \label{fig:3}
\end{figure}

The flow of the proposed method is shown in Figure \ref{fig:3}. The proposed method estimates the state (position, attitude, articulated angle) of a dump truck from four GNSSs by Graph-based optimization. Four "factors" of Graph-based optimization are used as constraints. We describe the detail in the next section.

\subsection{Factor Graph Structure}
Factor graphs have been used for various state estimation problems, mainly in the field of robotics \cite{go1}. In a factor graph, the state to be estimated is called a node, and a node represents state variable $X$. Multiple nodes at different times are connected at the edges via factor nodes. These edges correspond to an observation $Z$. The error function $e(\cdot)$ can be defined by the edges based on the node $X$ and observation Z. The state estimation problem can be expressed as a problem of minimizing the sum of all error functions as follows.

\begin{equation}
  \widehat{{X}}=\underset{{X}}{\operatorname{argmin}} \sum_{i}\left\|{e}\left({X}_{i}, {Z}_{i}\right)\right\|_{\Omega_{i}}^{2}
\end{equation}

\noindent
where $\Omega_{i}$ is the information matrix, which is the inverse of the variance of a observation $Z$. The information matrix represents the weight of an error function. 

Factors and edges are added to the graph structure as constraints on node X from various observations, and in general the final graph structure will be very complex. A nonlinear least-squares method is generally used to find the node $X$ that minimizes the sum of the error functions. In this study, we estimate the state of the dump truck robustly and accurately by creating various edges based on GNSS observations such as GNSS Doppler, pseudorange, and carrier-phase measurements. The node $X_i$  estimated in epoch $i$ is the 3D position of each GNSS antenna.

\begin{equation}
  {X}_{i}=\left[\begin{array}{llll}{X_{i}^\mathrm{1}} & {X_{i}^\mathrm{2}} & {X_{i}^\mathrm{3}} & {X_{i}^\mathrm{4}}\end{array}\right]^{T}=\left[\begin{array}{llllllllllll}{x_{i}^\mathrm{1}} & {y_{i}^\mathrm{1}} & {z_{i}^\mathrm{1}} & {x_{i}^\mathrm{2}} & {y_{i}^\mathrm{2}} & {z_{i}^\mathrm{2}} & {x_{i}^\mathrm{3}} & {y_{i}^\mathrm{3}} & {z_{i}^\mathrm{3}} & {x_{i}^\mathrm{4}} & {y_{i}^\mathrm{4}} & {z_{i}^\mathrm{4}}\end{array}\right]^{T}
\label{eq:state}
\end{equation}

\noindent
where the superscript represents the GNSS antenna number. The coordinate origin of the dump truck is defined as between antennas 1 and 2. The position of the dump truck $P_i$ is computed as follows:

\begin{equation}
  {P}_{i} = \frac{1}{2} \left({X}_{i}^\mathrm{1} + {X}_{i}^\mathrm{2}\right)
\label{eq:pos}
\end{equation}

The articulated angle  $\theta_i$ is computed from the difference between the orientation angles of the front and rear sections:

\begin{equation}
  {\theta}_{i} = {\theta}_{i}^\mathrm{12} - {\theta}_{i}^\mathrm{34}  
   =\tan ^{-1}\left( \dfrac {y_i^{2}-y_i^{1}}{x_i^{2}-x_i^{1}}\right)
   -\tan ^{-1}\left( \dfrac {y_i^{4}-y_i^{3}}{x_i^{4}-x_i^{3}}\right)
\label{eq:theta}
\end{equation}

\begin{figure}[t]
  \centering
  \includegraphics[width=150mm]{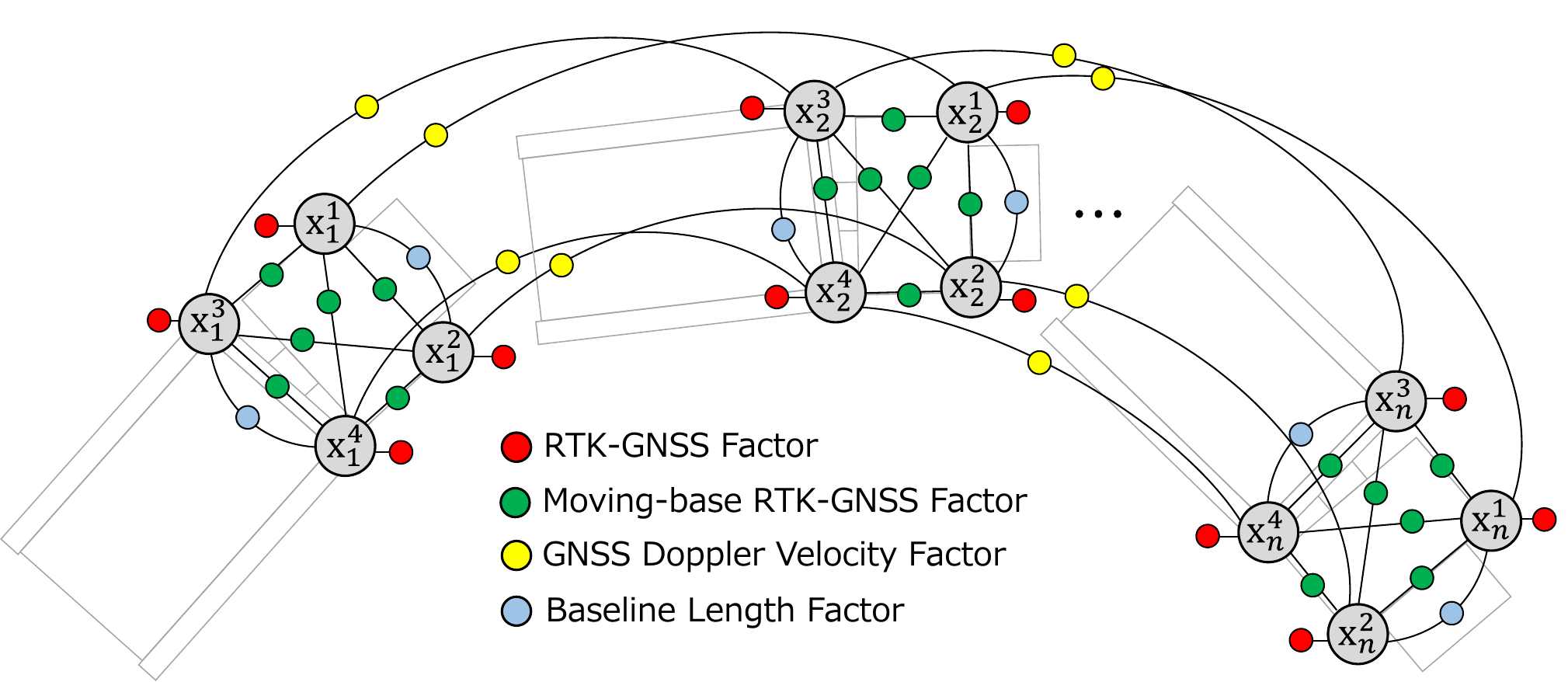} 
  \caption{Graphical structure of the proposed method. The color circles indicate the factor. RTK-GNSS, moving-base RTK-GNSS, GNSS Doppler velocity, and baseline length factors are used to estimate dump truck states.}
  \label{fig:4}
\end{figure}

Figure \ref{fig:4} shows the structure of graph-based optimization using the proposed method. The colored circles in Figure \ref{fig:4} represent the four main factors: RTK-GNSS, moving-base RTK-GNSS, GNSS Doppler velocity, and baseline length.

\subsection{RTK-GNSS Factor}
We used the RTK-GNSS factor (red circle in Figure \ref{fig:4}) to directly constrain the 3D position of the antenna. RTK-GNSS is a technique for accurately estimating 3D vectors from a base station whose position is known and fixed in an environment by taking the difference between the GNSS carrier-phase measurements between a base station and a GNSS receiver mounted on a mobile vehicle. GNSS carrier-phase measurements contain the integer ambiguity term, and when this ambiguity is correctly estimated, the 3D vectors can be estimated with centimeter accuracy. This accurate positioning solution is called the fixed solution. The observed carrier phase $\Phi_{i}^{k}$ from GNSS satellite $k$ at time $i$ is denoted as:

\begin{equation}
  \Phi_{i}^{k}=r_{i}^{k}+\lambda N_{i}^{k}+c\left(\delta t_{i}-\delta t_{i}^{k}\right)-I_{i}^{k}+T_{i}^{k}+\epsilon_{i}^{k}
\label{eq:cp}
\end{equation}

\noindent
where $r_{i}^{k}$ is the geometric distance between satellite $k$ and receiver at time $i$, $\lambda$ is the wavelength of the GNSS signal, $N_{i}^{k}$ is the integer ambiguity of the carrier phase, $c$ is the speed of light, $\delta t_{i}$ and $\delta t_{i}^{k}$ are receiver and satellite clock errors, respectively; $I_{i}^{k}$ and $T_{i}^{k}$ are ionospheric and tropospheric delay errors, respectively; and $\epsilon_{i}^{k}$ is receiver noise. In RTK-GNSS, two GNSS receivers are used to eliminate common errors in GNSS observations by subtracting two carrier-phase observations. Here, the ionospheric delay $I_{i}^{k}$, tropospheric delay $T_{i}^{k}$, and satellite clock error $\delta t^{k}$ are considered to be common errors at the same time and place, and can be eliminated by taking the difference between the observations from the two receivers. To eliminate the receiver clock error $\delta t_{i}$, we calculate the inter-satellite difference between satellites $k$ and $l$ of the inter-base station difference. This process is called double-difference (DD). The DD carrier-phase observation is

\begin{equation}
  \begin{aligned} \nabla \Delta \Phi_{i}^{k, l} &=\Delta \Phi_{i}^{k}-\Delta \Phi_{i}^{l} \\ &=\nabla \Delta r_{i}^{k, l}+\lambda \nabla \Delta N_{i}^{k,l}+\nabla \Delta \epsilon_{i}^{k,l} \end{aligned}
  \label{eq:DD}
  \end{equation}

\noindent
where $\Delta$ indicates differencing the measurements between the base and mobile stations, and $\nabla \Delta$ indicates differencing the measurements between the two satellites $k$ and $l$. Only the DD geometric distance $\nabla \Delta r_{i}^{k,l}$ and the DD carrier phase ambiguity $\nabla \Delta N_{i}^{k,l}$, which is an integer value, remain in this DD carrier phase observation. If the DD carrier phase ambiguity $\nabla \Delta N_{i}^{k,l}$ is correctly estimated, the 3D vector from the reference station ${B}_{i}=\left[\begin{array}{lll}{b_{x, i}} & {b_{y, i}} & {b_{z, i}}\end{array}\right]^{T}$ can be converted from the DD geometric distance $\nabla \Delta r_{i}^{k,l}$. We estimate the integer DD carrier phase ambiguity using the LAMBDA method \cite{lambda}, which is an integer least-squares method. Here, the ratio test \cite{gnss} is commonly used to check whether the integer ambiguity is correctly estimated, and the RTK-GNSS solution passing the ratio test is called as fix solutions. Only if the ratio test is passed is the 3D vector from the reference station added as a constraint. The error function is as follows:

\begin{equation}
  \boldsymbol{e}_{\mathrm{rtk},i}^{j}=X_i^j-B_i^j
  \label{eq:10}
  \end{equation}
  
  \begin{equation}
    \left\|e_{\mathrm{rtk},i}^{j}\right\|_{\Omega_{\mathrm{rtk}}}=e_{\mathrm{rtk},i}^{j} \Omega_\mathrm{rtk} e_{\mathrm{rtk},i}^{j}
  \label{eq:11}
  \end{equation}

\noindent
where $j$ is the antenna number. $\Omega_\mathrm{rtk}$ is the information matrix, which is the inverse of the RTK-GNSS variance, and is set to 0.1 m. If the GNSS satellites are not blocked by objects and the number of satellites is sufficiently large, the fixed solution can be calculated. However, the fix rate of RTK-GNSS decreases in an environment where satellites are blocked, such as in mountainous areas, owing to the decrease in the number of satellites, multipath, and other factors. In this study, we use not only the RTK-GNSS constraint but also the multiple moving-base RTK-GNSS constraints between the antennas and baseline length constraints.

\subsection{Moving-Base RTK-GNSS Factor}
In normal RTK-GNSS, the base station is fixed to the environment. In the moving-base RTK-GNSS, the base station moves \cite{movingbase}. Except that the base station is moving, the moving-base RTK-GNSS process is almost the same as that of normal RTK-GNSS, and the 3D vector from the moving base station is estimated by calculating the DD GNSS carrier phase as shown in equation (\ref{eq:DD}). This means that a highly accurate relative position vector among the antennas can be obtained by calculating the DD among multiple GNSS antennas installed on a dump truck. The green circles in Figure \ref{fig:4} indicate the moving-base RTK-GNSS factors. These 3D vectors were added to the graph as a geometric constraint between the antennas.

The number of combinations of four antennas without repetition is ${}_4 C _2=6$. We enumerate all antenna combinations and compute the moving-base RTK-GNSS solution. When the fix solution can be obtained, a moving-base RTK-GNSS constraint is added. The error function of the moving-base RTK-GNSS edge is expressed as follows.

\begin{equation}
  \boldsymbol{e}_{\mathrm{mvrtk},i}^{m}=X_i^{S^1(m)}-X_i^{S^2(m)}-B_i^m
  \label{eq:10}
  \end{equation}
  
  \begin{equation}
    \left\|e_{\mathrm{mvrtk},i}^{m}\right\|_{\Omega_{\mathrm{mvrtk}}}=e_{\mathrm{mvrtk},i}^{m} \Omega_\mathrm{mvrtk} e_{\mathrm{mvrtk},i}^{m}
  \label{eq:11}
  \end{equation}

\noindent
where $S^1(m)$ and $S^2(m)$ represent the antenna number of the $m$-th GNSS antenna combinations. $B_i^m$ is the 3D baseline vector between the antennas of the $m$-th GNSS antenna combinations, which is computed from the moving-base RTK-GNSS. $\Omega_\mathrm{mvrtk}$ is an information matrix that is inverse of the variance of moving-base RTK-GNSS, and it is set to 0.1 m as with normal RTK-GNSS. Similar to the RTK-GNSS factor, the moving-base RTK-GNSS factor is added as a constraint only when a fixed solution is obtained. The four antennas provide redundancy, and the moving-base RTK-GNSS factor can improve the robustness in environments where the GNSS positioning accuracy decreases.

\subsection{GNSS Doppler Velocity Factor}
The edge between the states owing to the Doppler velocity was used in this study. The relative velocity between the GNSS antenna and the GNSS satellite can be calculated by measuring the Doppler shift of the satellite’s signal using a single GNSS receiver. The exact velocity of the GNSS satellite can be calculated from the satellite orbit information transmitted by the GNSS satellite. As a result, the accurate 3D velocity of the GNSS antenna can be estimated from the observed GNSS Doppler shift from multiple satellites. We use this velocity as a constraint between nodes. The accuracy of the Doppler velocity can be estimated within cm/s under open-sky conditions \cite{vel1}.

The 3D velocity estimated from the Doppler at time $i$ is denoted as ${V}_{i}=\left[\begin{array}{lll}{v_{x, i}} & {v_{y, i}} & {v_{z, i}}\end{array}\right]^{T}$. $\Delta t$ is the time step of the GNSS observation. The error function of Doppler velocity constraint is

\begin{equation}
{e}_{\mathrm{vel},i}^j=\left(X_{i}^j-X_{i-1}^j\right)-V_{i}^j \cdot \Delta t
\label{eq:6}
\end{equation}

\begin{equation}
  \left\|e_{\mathrm{vel},i}^{j}\right\|_{\Omega_{\mathrm{vel}}}=e_{\mathrm{vel},i}^{j} \Omega_\mathrm{vel} e_{\mathrm{vel},i}^{j}
\label{eq:11}
\end{equation}

\noindent
where $\Omega_\mathrm{vel}$ is the information matrix of the Doppler velocity. The variance of the Doppler velocity is set to 0.1 m/s in this study.

\subsection{Baseline Length Factor}
We use the geometric constraint between the GNSS antennas to estimate the position and articulated angle of the dump truck. The relative positions of the two antennas mounted on the front and rear sections of the dump truck were considered to be invariant. The 3D distances between the two antennas are denoted as $L_{12}$ and $L_{34}$, and must be constant value. The distance between each antenna is measured in advance. The error function of the geometric constraint between the antennas is as follows:

\begin{equation}
  e_{\mathrm{ant},i}=\left\|X_i^\mathrm{2}-X_i^\mathrm{1}\right\|-L_{12}+\left\|X_i^\mathrm{4}-X_i^\mathrm{3}\right\|-L_{34}
\label{eq:4}
\end{equation}

\begin{equation}
  \left\|e_{\mathrm{ant},i}\right\|_{\Omega_{\mathrm{ant}}}=e_{\mathrm{ant},i} \Omega_\mathrm{ant} e_{\mathrm{ant},i}
\label{eq:5}
\end{equation}

\noindent
where $\Omega_{\mathrm{ant}}$ is the information matrix of the inter-antenna geometric constraint and is the inverse of the variance of the inter-antenna distance. The distance between the antennas is not changed, therefore, this variance is set to a very small value, and a variance of 0.01 m was used in this study.

\subsection{Optimization}
By considering all the factors, we can finally obtain the following objective function to be minimized.

\begin{equation}
  \begin{aligned} 
  \widehat{\boldsymbol{X}}=\underset{X}{\operatorname{argmin}} 
  \sum_{i} \sum_{j}\left\|e_{\mathrm{rtk}, i}^{j}\right\|_{\Omega_{\mathrm{rtk}}}^{2} +
  \sum_{i} \sum_{m}\left\|e_{\mathrm{mvrtk}, i}^{m}\right\|_{\Omega_{\mathrm{mvrtk}}}^{2} +
  \sum_{i} \sum_{j} \left\|e_{\mathrm{vel}, i}^{j}\right\|_{\Omega_{\mathrm{vel}}}^{2}+
  \sum_{i}\left\|e_{\mathrm{ant}, i}\right\|_{\Omega_{ \mathrm{ant}}}^{2}
  \end{aligned}
  \label{eq:all}
  \end{equation}

Various optimization methods can be used to find the optimal solution of this equation, and in this study, we used the dogleg optimizer \cite{tuto}. As a result, the 3D positions of each of the four GNSS antennas are obtained, and finally, the dump truck position and articulated angle are calculated from equations (\ref{eq:pos}) and (\ref{eq:theta}). The first, second, and third terms in the equation (\ref{eq:all}) are generated from GNSS observations, while the fourth term is due to the geometric arrangement of the antenna. 

For the RTK-GNSS and Moving-base RTK-GNSS factors, we only add a factor when we have a reliable fix solution that solves the carrier phase ambiguity. The fixed baseline length factor is always a valid factor and is the most reliable. By integrating and using multiple factors, the proposed method can robustly estimate the state of the dump truck. In addition, the proposed method does not require any sensor other than GNSS and can be easily used.

\section{Experiments}
\subsection{Setup}
\begin{figure}[t]
  \centering
  \includegraphics[width=125mm]{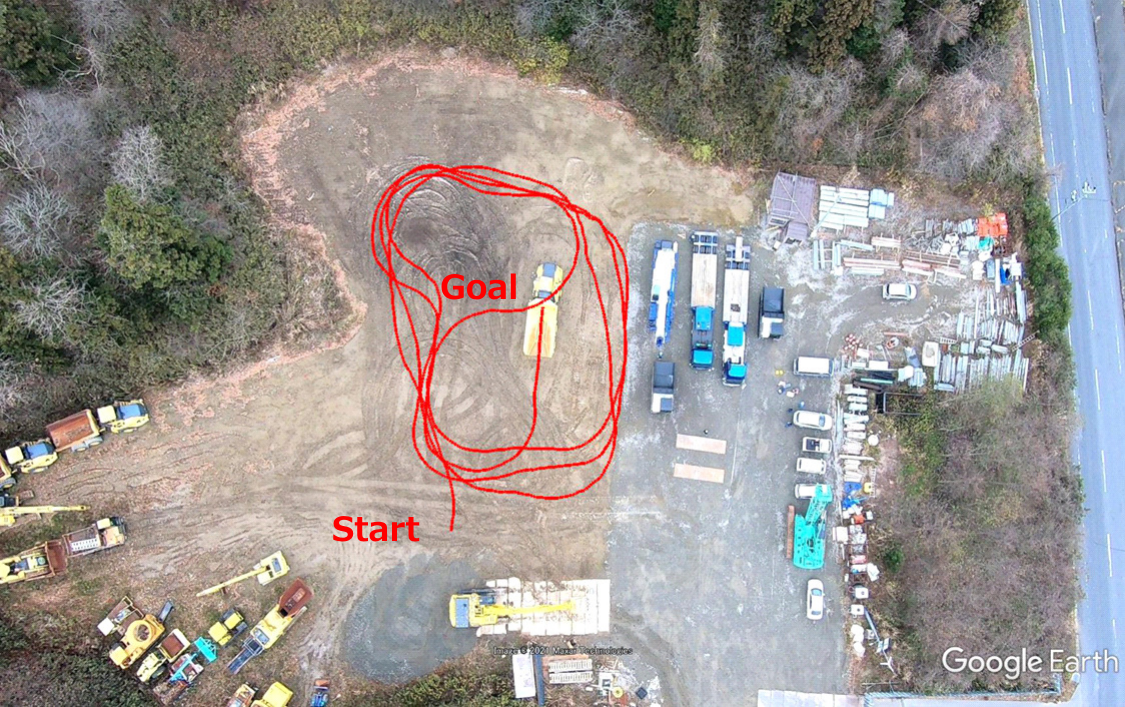} 
  \caption{Aerial photograph of the experimental field. The red line indicates the travel path of the articulated dump truck. The experimental field is an open-sky environment with few obstacles blocking the GNSS signals. }
  \label{fig:5}
\end{figure}

To evaluate the proposed method, a driving experiment was conducted on an articulated dump truck in an actual field. As shown in Figure \ref{fig:2}, an articulated dump truck (Komatsu Ltd., HM400-5) was equipped with four u-blox NEO-M8T GNSS receivers and four Tallysman TW3742 antennas. These GNSS receivers are low-cost single-frequency GNSS receivers. GNSS data were collected at a rate of 10 Hz. We used GPS, BeiDou, Galileo, and QZSS for the RTK-GNSS computation. GNSS base station was set up at close distances in the experimental field and were used for RTK-GNSS. Figure \ref{fig:5} shows an aerial photograph of the experimental field and the travel path of the articulated dump truck (red line). The traveling environment was flat and there are no objects that block GNSS signals. As shown in Figure \ref{fig:5} , the dump truck made repeated left and right turns at a speed of approximately 10 km/h. For the performance evaluation, a position and attitude estimation system (NovAtel SPAN-CPT7) \cite{span} was used to estimate the reference positions as a ground truth. This system integrates high-grade INS and a multiple-frequency GNSS receiver, and this sensor provides position and attitude with sufficient accuracy for reference. RTKLIB \cite{rtklib}, an open-source GNSS data processing library, was used for GNSS calculations, such as RTK-GNSS. GTSAM \cite{gtsam} was used at the backend for the factor graph optimization. The performance evaluation of the proposed method was performed during the post-processing.

\subsection{Evaluation Method}
As shown in Figure \ref{fig:5} , the driving environment is an open-sky environment, but actual construction sites are often located in mountainous areas, and there is a concern that GNSS satellites may be shielded and GNSS positioning performance may be degraded. In this study, we evaluate the robustness and accuracy of the proposed method by simulating the shielding  of GNSS satellites by adding a pseudo GNSS satellite elevation masking process. Specifically, we evaluated the proposed method under three conditions: (a) no satellite elevation mask, (b) satellite elevation mask: 35$^\circ$, and (c) satellite elevation mask: 45$^\circ$. The proposed method is compared with the conventional RTK-GNSS method for dump truck position estimation \cite{our1} and with the LiDAR method for articulated angle estimation. 

\begin{figure}[t]
  \centering
  \includegraphics[width=150mm]{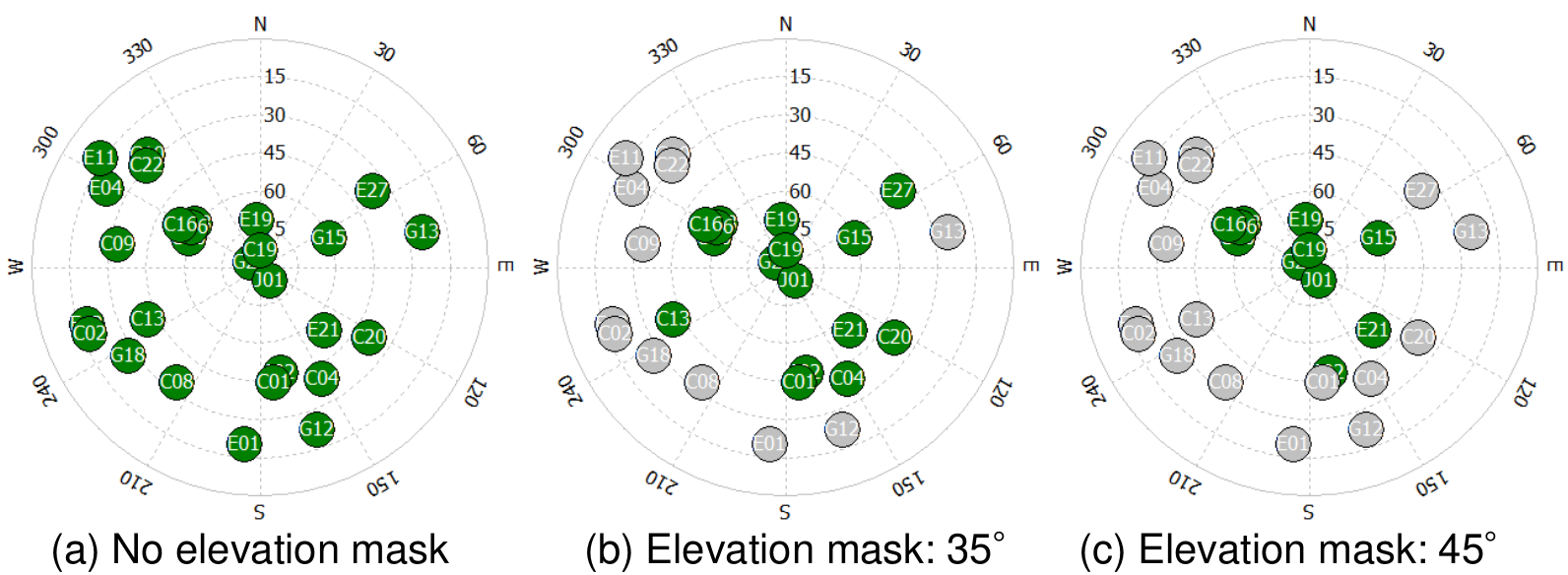} 
  \caption{GNSS satellite constellation during the experiment. We evaluate the proposed method on three different GNSS elevation masks (no mask, 35$^\circ$, and 45$^\circ$). In the figure, "G" represents GPS satellites, "C" represents BeiDou, "E" represents Galileo, and "Q" represents QZSS satellites.}
  \label{fig:6}
\end{figure}

\begin{figure}[t]
  \centering
  \includegraphics[width=140mm]{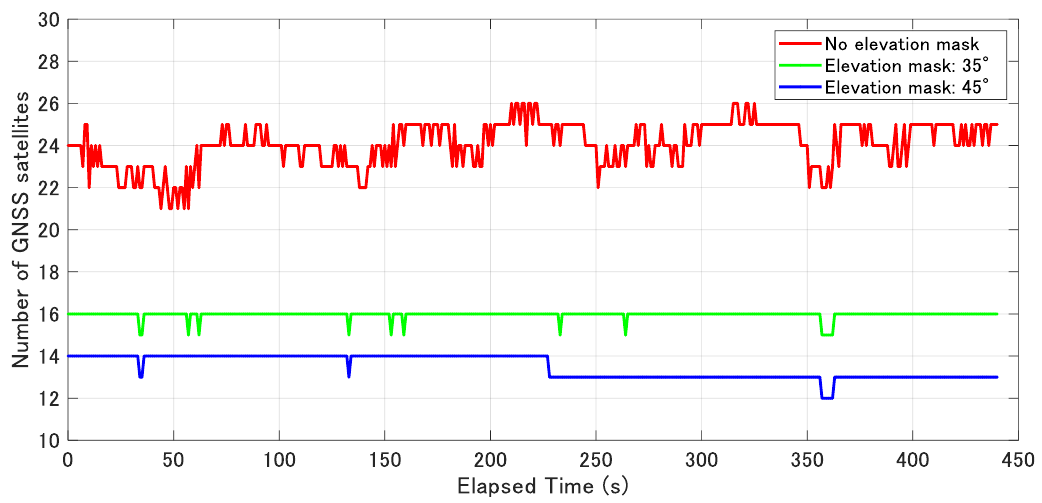} 
  \caption{Change in the number of available satellites when using different satellite elevation masks. The total number of available satellites was reduced by applying an elevation mask to the GNSS satellites to simulate a mountainous environment, eliminating satellites with low elevation angles.}
  \label{fig:7}
\end{figure}

\subsection{Satellite Constellation}
Figure \ref{fig:6}  shows the constellation of GNSS satellites during the experiment. G, J, E, and C in Figure \ref{fig:6} indicate GPS, QZSS, Galileo, and BeiDou, respectively. Figures \ref{fig:6}(a), \ref{fig:6}(b), and \ref{fig:6}(c) show the configuration of the satellites when the elevation mask is applied. The gray satellites in Figure \ref{fig:6} were excluded by the elevation mask and were not used for GNSS positioning. Figure \ref{fig:7} shows the time series of the number of GNSS satellites in each elevation mask. The average number of available satellites is 24 for the no-elevation mask, 16 for the 35$^\circ$ elevation mask, and 13 for the 45$^\circ$ elevation mask. In an actual mountainous area, the number of available satellites is reduced, which may adversely affect the GNSS positioning accuracy.

\begin{figure}[t]
  \centering
  \includegraphics[width=130mm]{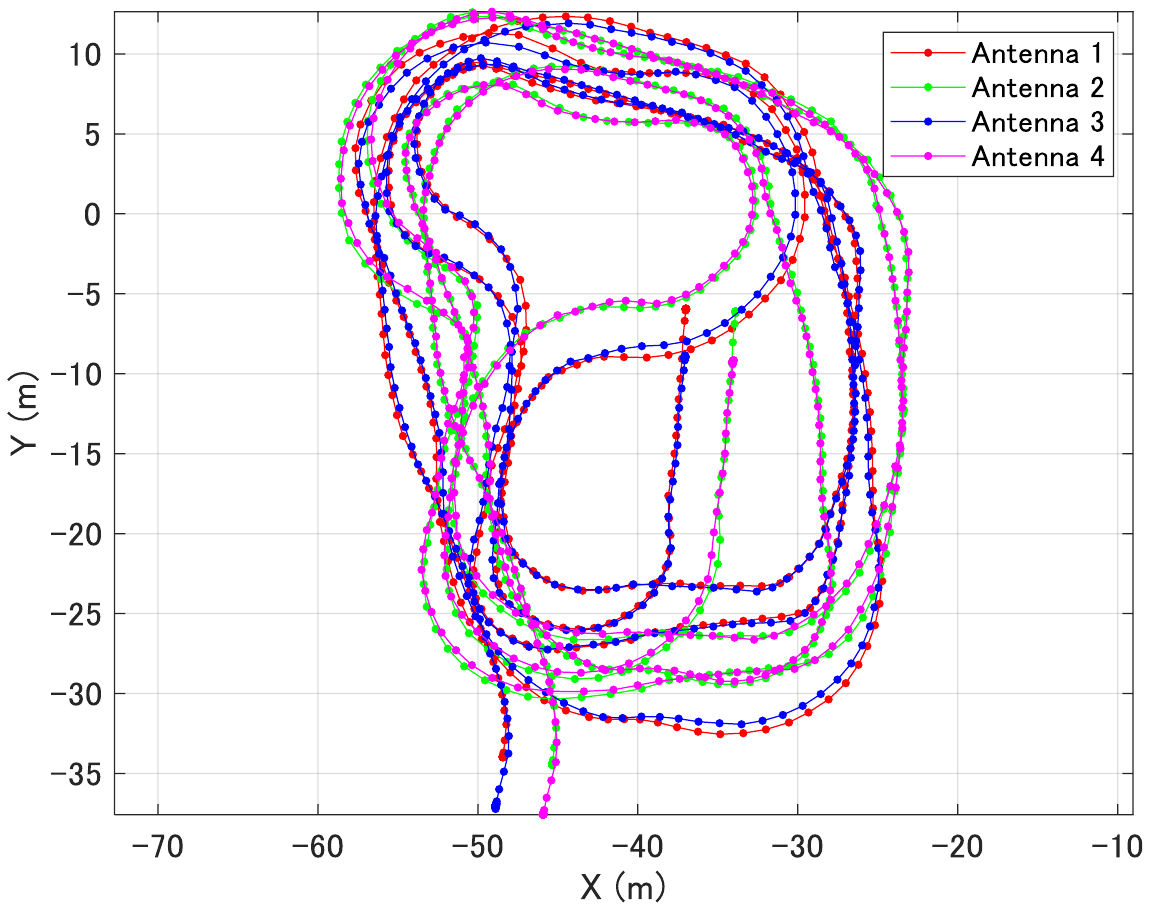} 
  \caption{Estimated GNSS antenna positions using the proposed method. The colors of the plots indicate the trajectory of each GNSS antenna. The trajectory of each antenna is smoothly estimated.}
  \label{fig:8}
\end{figure}

\begin{figure}[t]
  \centering
  \includegraphics[width=140mm]{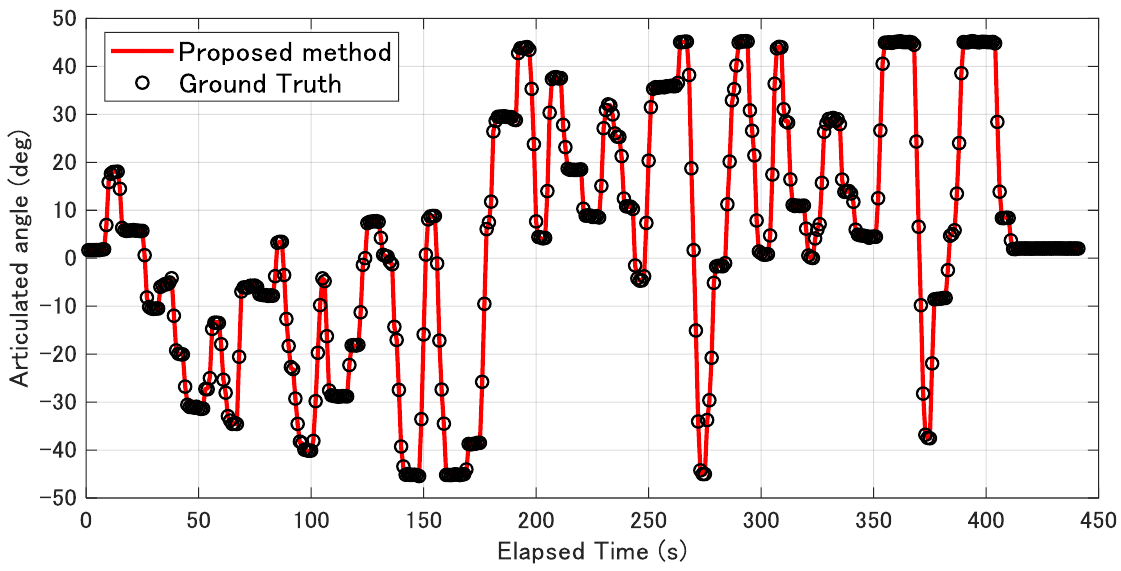} 
  \caption{Estimated articulated angle using the proposed method. The red line indicates the proposed method and the black circles are the ground truth.}
  \label{fig:9}
\end{figure}

\subsection{Estimation Results}
Figure \ref{fig:8} shows the positions of the four GNSS antennas on the dump truck estimated using the proposed method (without an elevation mask). From the 3D positions of each antenna, the dump truck position and articulated angle are calculated using Equations (\ref{eq:pos}) and (\ref{eq:theta}). Figure \ref{fig:9} shows the time-series variation of the estimated articulated angle (no elevation mask). As shown in this figure, it can be confirmed that the articulated angle changes significantly in accordance with the left and right turns.

\begin{figure}[t]
  \centering
  \includegraphics[width=160mm]{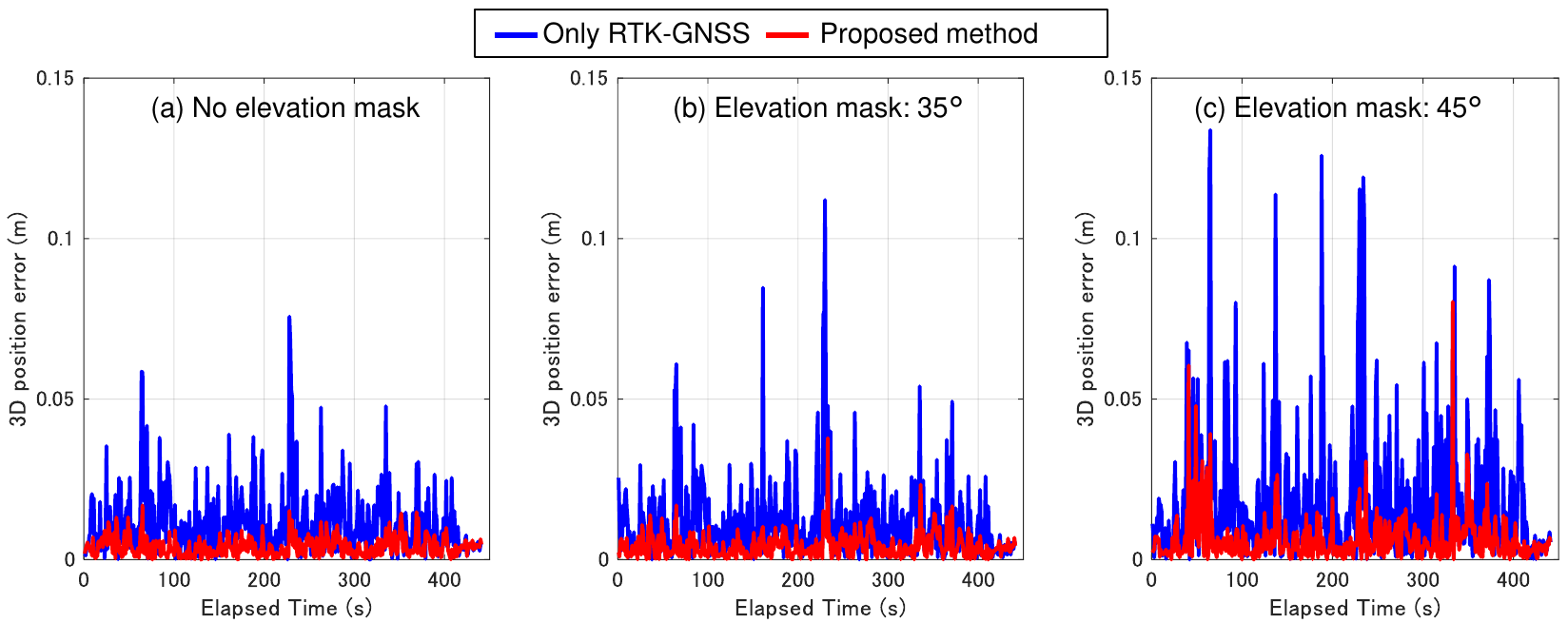} 
  \caption{Comparison of 3D localization error. The red and blue lines indicate the proposed method and the method uses only RTK-GNSS.}
  \label{fig:10}
\end{figure}

Figure \ref{fig:10} and Table 1 show the 3D root-mean-square (RMS) error of the dump truck position compared to the ground truth in each elevation mask. From these figures, it can be confirmed that the fix rate of RTK-GNSS decreases as the elevation mask increases in the method using normal RTK-GNSS, and the position estimation accuracy deteriorates slightly. On the other hand, the proposed method maintains a position estimation accuracy of approximately 3 cm even when the elevation mask is 45$^\circ$, indicating that the proposed method can accurately estimate the dump track position in harsh environments.

\begin{figure}[t!]
  \centering
  \includegraphics[width=160mm]{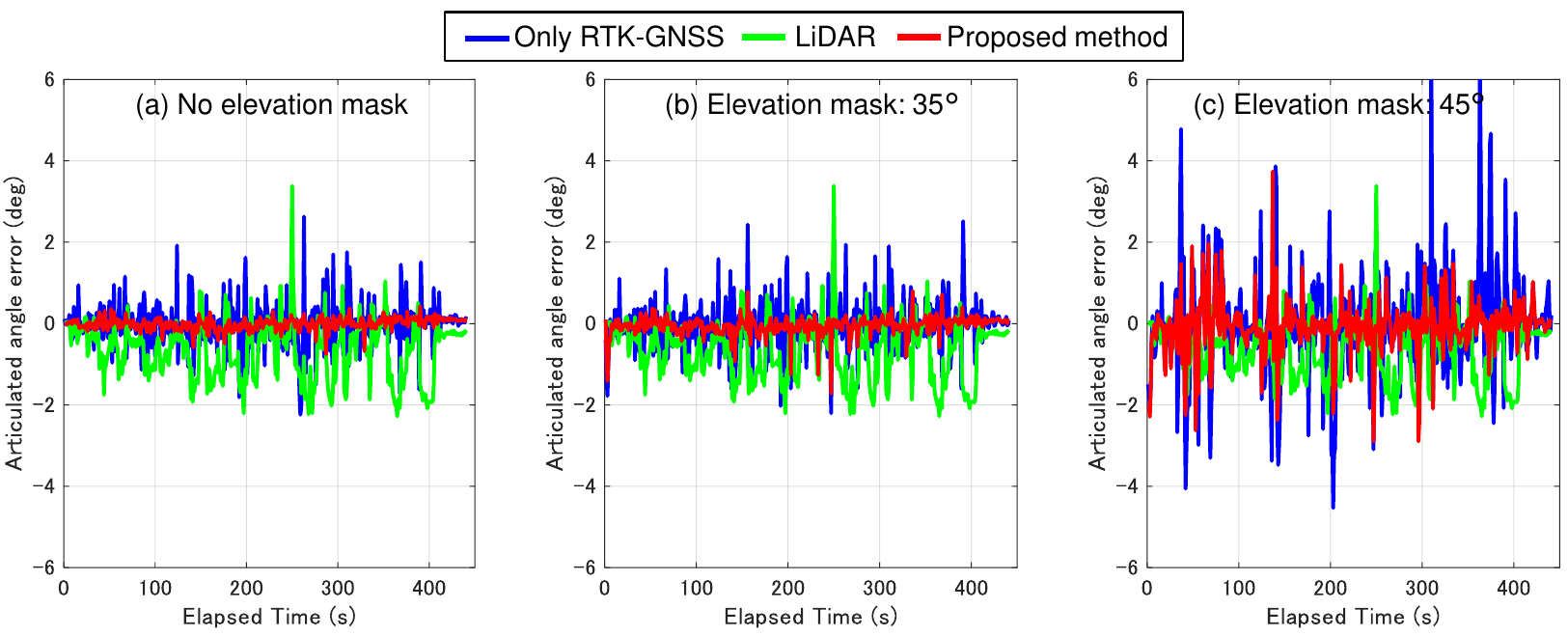} 
  \caption{Comparison of articulated angle error. The red, blue, and green lines indicate the proposed method, the method uses only RTK-GNSS, and the method uses LiDAR, respectively.}
  \label{fig:11}
\end{figure}

\begin{table}[t]
  \tbl{Comparison of 3D RMS localization error at each elevation mask angle. As the satellite elevation angle mask is increased, the estimation accuracy decreases, but the proposed method shows the best estimation accuracy.}
  {\begin{tabular}{c|ccc}
    \hline
                    & No elevation mask & Elevation mask: 35$^\circ$ & Elevation mask: 45$^\circ$ \\ \hline
    Only RTK-GNSS   & 0.028 m           & 0.031 m             & 0.062 m             \\
    Proposed method & 0.021 m           & 0.021 m             & 0.031 m             \\ \hline
    \end{tabular}}
  \label{table:1}
\end{table}

\begin{table}[t!]
  \tbl{Comparison of RMS error of articulated angle. Even when the satellite elevation mask is applied, the proposed method can estimate the articulated angle more accurately than the conventional method using LiDAR.}
  {\begin{tabular}{c|ccc}
    \hline
    \multicolumn{1}{l|}{} & No elevation mask & Elevation mask: 35$^\circ$ & Elevation mask: 45$^\circ$ \\ \hline
    Only RTK-GNSS         & 0.490$^\circ$     & 0.548$^\circ$              & 1.588$^\circ$              \\
    Proposed method       & 0.132$^\circ$     & 0.215$^\circ$              & 0.766$^\circ$              \\ \hline
    LiDAR                 & \multicolumn{3}{c|}{0.978$^\circ$}                                          \\ \hline
    \end{tabular}}
  \label{table:2}
\end{table}

Figure \ref{fig:11} and Table 2 show the RMS error of the articulated angle compared to ground truth. Figure \ref{fig:11} shows that the value of error by the method using LiDAR increased at the point where the articulated angle became larger. The proposed method can measure angles with higher accuracy (0.1$^\circ$) than the RTK-GNSS and LiDAR methods in an open-sky environment. The proposed method can also estimate the articulated angle with an accuracy of 0.76$^\circ$ even when the elevation mask is 45$^\circ$. This means that the proposed method can measure the articulated angle more accurately than the LiDAR method even in a mountainous environment.

\subsection{Discussion}
It is confirmed that the proposed method can estimate the position and articulated angle of the dump trucks with higher accuracy than the conventional method in an open-sky environment. When the number of available satellites is limited while simulating a mountainous area, the estimation accuracy of the proposed method is maintained, while the accuracy of the conventional method using only RTK-GNSS is greatly reduced. This is because the geometric constraint between GNSS antennas, which utilizes the redundancy of multiple antennas, works effectively. In addition, the proposed method can be used by simply installing low-cost GNSS antennas on the front and rear sections of the articulated dump trucks and does not require a complicated calibration process as in the case of LiDAR. The proposed method will greatly facilitate the widespread use of automatic dump truck operations because of its low cost and ease of utilization. In summary, the effectiveness of the proposed method for estimating the state of articulated dump trucks was confirmed.

\section{Conclusion}
We proposed a method to estimate the position and articulation angle of an articulated six-wheeled dump truck with high accuracy and robustness by using multiple GNSS antennas. Four GNSS antennas are mounted on the dump truck, and the constraint between multiple GNSS antennas by RTK-GNSS and the constraint of the geometric arrangement of antennas are proposed. Additionally, we proposed a method to estimate the state of a dump truck using graph optimization. The accuracy and robustness of the proposed method are evaluated using field tests. In an experiment using an elevation mask that simulates the shielding of a GNSS satellite in a mountainous area, it was confirmed that the proposed method could estimate the position and articulated angle more accurately than the conventional method.

In future studies, we will combine the proposed method with INS. In combination with INS, the state estimation of a dump truck will be more robust in environments where GNSS signals are more blocked. In this study, we evaluated the accuracy of the proposed method using post-processing. We also plan to evaluate the real-time estimation of the position and articulated angle of the proposed method using the iSAM \cite{isam} framework.

\section*{Acknowledgement}
This research was supported by NEDO's commissioned research project (No. 18065741).

\bibliographystyle{tADR}
\bibliography{ar2021}

\end{document}